\documentclass[lettersize,journal]{IEEEtran}
\usepackage{amsmath,amsfonts}
\usepackage[ruled]{algorithm2e}
\usepackage{algorithmic}
\usepackage{array}
\usepackage[caption=false,font=normalsize,labelfont=sf,textfont=sf]{subfig}
\usepackage{textcomp}
\usepackage{stfloats}
\usepackage{url}
\usepackage{verbatim}
\usepackage{graphicx}
\usepackage{booktabs}
\usepackage{multirow}
\usepackage{makecell}
\usepackage{amssymb}
\usepackage{nomencl} 
\usepackage[colorlinks,linkcolor=blue]{hyperref}
\makenomenclature

\usepackage{etoolbox}
\renewcommand\nomgroup[1]{%
  \item[\bfseries
  \ifstrequal{#1}{A}{Sets}{%
  \ifstrequal{#1}{B}{Parameters}{%
  \ifstrequal{#1}{C}{Decision
Variables}{}}}%
]}

\def\BibTeX{{\rm B\kern-.05em{\sc i\kern-.025em b}\kern-.08em
    T\kern-.1667em\lower.7ex\hbox{E}\kern-.125emX}}
\usepackage{balance}

\begin{document}

\title{Deep Reinforcement Learning for Solving the Fleet Size and Mix Vehicle Routing Problem}

\author{Pengfu~Wan, Jiawei~Chen, and Gangyan~Xu 
\thanks{This work has been submitted to the IEEE for possible publication. Copyright may be transferred without notice, after which this version may no longer be accessible.

Pengfu Wan, Jiawei Chen, and Gangyan Xu are with the Department of Aeronautical and Aviation Engineering, The Hong Kong Polytechnic University, Hung Hom, Kowloon, Hong Kong. Email: pengfu.wan@connect.polyu.hk, superlaser-jw.chen@connect.polyu.hk, gangyan.xu@polyu.edu.hk.}
}


\maketitle

\begin{abstract}
The Fleet Size and Mix Vehicle Routing Problem (FSMVRP) is a prominent variant of the Vehicle Routing Problem (VRP), extensively studied in operations research and computational science. FSMVRP requires simultaneous decisions on fleet composition and routing, making it highly applicable to real-world scenarios such as short-term vehicle rental and on-demand logistics. However, these requirements also increase the complexity of FSMVRP, posing significant challenges, particularly in large-scale and time-constrained environments.
In this paper, we propose a deep reinforcement learning (DRL)-based approach for solving FSMVRP, capable of generating near-optimal solutions within a few seconds. Specifically, we formulate the problem as a Markov Decision Process (MDP) and develop a novel policy network, termed FRIPN, that seamlessly integrates fleet composition and routing decisions. Our method incorporates specialized input embeddings designed for distinct decision objectives, including a remaining graph embedding to facilitate effective vehicle employment decisions.
Comprehensive experiments are conducted on both randomly generated instances and benchmark datasets. The experimental results demonstrate that our method exhibits notable advantages in terms of computational efficiency and scalability, particularly in large-scale and time-constrained scenarios. These strengths highlight the potential of our approach for practical applications and provide valuable inspiration for extending DRL-based techniques to other variants of VRP.
\end{abstract}


\printnomenclature

\nomenclature[A, 01]{\( \mathcal{N} \)}{Set of costumers, $\mathcal{N} = \{1, 2, \dots, n\}$.}
\nomenclature[A, 02]{\( \mathcal{V} \)}{Set comprising the depot and a set of customers, $\mathcal{V} = \{0, 1, 2, \dots, n\}$.}
\nomenclature[A, 03]{\( \mathcal{E} \)}{Set of edges.}
\nomenclature[A, 04]{\( \mathcal{G} \)}{Graph set, $\mathcal{G} = (\mathcal{V}, \mathcal{E})$.}
\nomenclature[A, 05]{\( \mathcal{K} \)}{Set of vehicle types.}

\nomenclature[B, 01]{\( f^{k} \)}{Fixed cost of vehicles of type $k$.}
\nomenclature[B, 02]{\( c^k \)}{Variable cost of vehicles of type $k$ per unit distance.}
\nomenclature[B, 03]{\( Q^k \)}{Capacity of vehicles of type $k$.}
\nomenclature[B, 04]{\( e_{ij} \)}{Travel distance between nodes $i$ and $j$.}
\nomenclature[B, 05]{\( d_{i} \)}{Demand of node $i$.}

\nomenclature[C, 01]{\(x_{ij}^k\)}{Binary decision variables, 1 if vehicle of type $k$ travels from node $i$ to $j$, 0 otherwise.}
\nomenclature[C, 02]{\(u_i^k\)}{Continuous variables, cumulative load delivered by a vehicle of type $k$ upon arrival at customer $i$}

\section{Introduction}
\IEEEPARstart{T}{he}
Fleet Size and Mix Vehicle Routing Problem (FSMVRP) represents a critical extension of the classical Capacitated Vehicle Routing Problem (CVRP), arising frequently in real-world logistics and transportation scenarios \cite{golden1984fleet, hiermann2016electric}. In many practical applications, such as short-term vehicle rentals or on-demand logistics services \cite{yang2008car, chen2025demand}, decision-makers are confronted with the challenge of composing an optimal fleet from a heterogeneous set of available vehicles, each characterized by distinct capacities, rental costs, and operational costs. Unlike another important routing problem, the Heterogeneous Capacitated Vehicle Routing Problem (HCVRP) \cite{Li2021DeepRL, feng2019solving}, where the fleet composition is fixed and homogeneous, the FSMVRP requires simultaneous optimization of both the routing of vehicles and the composition of fleet size and mix. This dual consideration introduces additional layers of complexity, as it necessitates balancing the variable costs incurred by route planning with the fixed costs associated with vehicle acquisition or rental. Consequently, the FSMVRP more accurately reflects the operational realities faced by modern logistics providers, where cost-effective and flexible fleet management is paramount.

To address the FSMVRP, a variety of solution methodologies have been proposed, broadly categorized into exact algorithms and heuristic approaches. Exact methods, such as branch-and-bound or branch-and-cut algorithms \cite{toth1998exact, lahyani2018alternative}, are capable of guaranteeing optimal solutions but are inherently limited to small-sized problem instances due to their exponential computational complexity and substantial time requirements. In contrast, heuristic and metaheuristic algorithms, including genetic algorithms (GA)\cite{liu2009effective}, tabu search (TS)\cite{brandao2009deterministic}, and variable neighborhood search (VNS)\cite{rezgui2019application}, have demonstrated greater scalability and are widely adopted for tackling larger and more complex FSMVRP instances. However, these approaches typically rely on iterative improvement processes to explore the solution space, which can still be computationally intensive and time-consuming, especially when high-quality solutions are sought. These limitations become particularly pronounced in time-sensitive and large-scale commercial scenarios, such as green and smart logistics \cite{wang2023partial, mehlawat2019hybrid} logistics, where rapid and efficient decision-making is crucial. As a result, there remains a pressing need for more efficient and adaptive solution frameworks that can effectively balance solution quality and computational efficiency in real-world FSMVRP applications.

In recent years, deep reinforcement learning (DRL) has emerged as a promising paradigm for rapidly generating high-quality feasible solutions to complex combinatorial optimization problems \cite{bello2016neural, bengio2021machine, shao2021multi, jiang2021learning, wang2023multiobjective}. Notably, DRL-based approaches have achieved remarkable success in solving the routing problems and their various extensions \cite{kool2018attention, xu2021reinforcement, chen2024cooperative}, demonstrating superior performance in terms of both solution quality and computational efficiency. These methods leverage the powerful representation and decision-making capabilities of deep neural networks to learn effective routing policies directly from data, enabling swift adaptation to diverse problem instances. However, existing DRL frameworks are predominantly designed for routing optimization given a fixed fleet composition, and thus are not readily applicable to scenarios where fleet size and mix must also be determined. Specifically, extending DRL to address the FSMVRP introduces several unique challenges as follows:

First, the FSMVRP introduces a significantly higher level of complexity compared to traditional VRP variants. While classical approaches typically focus solely on optimizing vehicle routes, the FSMVRP requires the simultaneous determination of both fleet composition and routing strategies \cite{golden1984fleet}, with both sets of decisions jointly influencing the objective function. Effectively modeling this dual-decision process within a sequential decision-making framework is non-trivial, as it demands the integration of heterogeneous decision variables and the design of mechanisms that can capture their intricate interdependencies.

Second, the heterogeneous nature of the decision variables in FSMVRP presents unique challenges in terms of decision timing and contextual awareness. Specifically, the composition of fleet size and mix has to be interleaved with the routing decisions, and distinct considerations govern each type of decision. In particular, fleet composition is driven by cost structures and capacity requirements, whereas routing optimization depends on spatial and demand distributions. Designing a DRL framework that can appropriately sequence and coordinate these heterogeneous decisions, while accounting for their respective influencing factors, is essential for achieving high-quality solutions.

Third, the requirement for policy generalization is substantially heightened in the context of FSMVRP. While DRL-based methods for routing problems are typically expected to generalize across varying customer locations and demands \cite{jiang2023ensemble, hu2025solving}, FSMVRP further necessitates robust generalization across diverse candidate fleets. In practical applications, the available fleet composition may vary significantly from instance to instance. Therefore, it is imperative that the learned DRL policy remains effective not only for different routing scenarios but also for a wide range of fleet configurations, ensuring broad applicability and adaptability in real-world logistics environments.

Taking the above challenges into consideration, this work develops a DRL-based method for the FSMVRP. The contributions of this work lie in the following four aspects:
\begin{itemize}
    \item We formulate the FSMVRP as a Markov Decision Process (MDP), capturing the sequential nature of both fleet composition and routing decisions. This modeling framework provides a principled foundation for leveraging reinforcement learning techniques to address the inherent complexity of the FSMVRP.
    \item We propose a novel policy network architecture, termed the Fleet-and-Route Integrated Policy Network (FRIPN), which seamlessly integrates fleet composition and routing optimization within a unified decision-making framework. The proposed architecture is inherently flexible and can be readily extended to accommodate scenarios with varying candidate vehicle types, thereby enhancing its applicability to diverse real-world settings.
    \item To effectively address the heterogeneous nature of the decision variables, we design specialized input embeddings tailored to the distinct requirements of fleet composition and routing decisions. Furthermore, recognizing the critical influence of remaining customer demand on fleet composition, we introduce the remaining graph embedding to guide the policy network in making more informed fleet composition choices.
    \item Extensive computational experiments demonstrate the superior performance of our approach. Comparative analyses against current methods confirm its outstanding solution quality, particularly under stringent time constraints and for large-scale problem instances. Additional evaluations on established benchmark datasets further validate the performance of the proposed method.
\end{itemize}

The rest of the paper is structured as follows. Related works are reviewed in Section \ref{rw}. The mathematical model of the FSMVRP and related Markov Decision Process (MDP) are presented in Section \ref{problem_statement}. Section \ref{method} discusses the detailed DRL-based solution method. Then experimental results are given in Section \ref{experiment}. Finally, Section \ref{conclusion} concludes the whole paper.

\section{Related Work} \label{rw}

In this section, we will initially examine existing optimization approaches for addressing the FSMVRP, followed by a summary of the DRL-based methods tailored for routing problems.

\subsection{Optimization Methods for FSMVRP}

The FSMVRP was first introduced by Golden et al. \cite{golden1984fleet}, marking a significant advancement in the study of vehicle routing problems by incorporating fleet composition decisions alongside traditional routing optimization. Since its inception, a variety of optimization methods have been developed to tackle FSMVRP, which can be broadly categorized into exact algorithms and heuristic approaches. Exact methods, such as linear programming solvers \cite{yaman2006formulations, baldacci2009unified} employing branch-and-bound \cite{toth1998exact} and branch-and-cut \cite{lahyani2018alternative} techniques, have been used to guarantee optimal solutions for small-scale instances of FSMVRP.

Despite their theoretical appeal, exact algorithms often require substantial computational resources and extended solution times, especially as problem size and complexity increase. Consequently, heuristic-based methods have become the predominant choice for solving FSMVRP instances, aiming to efficiently generate high-quality feasible solutions within reasonable time frames. For example, Brandão \cite{brandao2009deterministic} proposed a heuristic algorithm based on TS to solve FSMVRP. Hiermann et al. \cite{hiermann2016electric} developed an adaptive large neighborhood search (ALNS) algorithm to deal with the electric fleet size and mix vehicle routing problem with time windows and recharging stations (E-FSMFTW), which considered the limited battery capacity and time windows at customer locations. Other heuristic-based algorithms, such as GA \cite{liu2009effective}, hybrid evolutionary algorithm \cite{kocc2014fleet}, and set partitioning-based heuristics \cite{salhi2013fleet} have also demonstrated considerable success in numerical experiments and practical applications.

In summary, while both exact and heuristic optimization methods have contributed significantly to the advancement of FSMVRP research, they remain constrained by computational efficiency and scalability issues. As problem size grows or real-time decision-making becomes necessary, these traditional approaches may struggle to deliver timely and effective solutions, highlighting the need for novel methodologies that can better address the demands of modern logistics environments.

\subsection{DRL for Routing Problems}

DRL has emerged as a powerful tool for addressing combinatorial optimization (CO) problems \cite{bello2016neural, barrett2020exploratory, berto2025rl4co}, owing to its ability to learn complex decision-making policies directly from data. Among the various CO problems, routing problems have attracted considerable attention due to their practical relevance and inherent NP-hardness challenges. Recent years have witnessed a surge of DRL-based approaches for solving classical routing problems, such as the traveling salesman problem (TSP)\cite{kool2018attention, xiao2024reinforcement}, the capacitated vehicle routing problem (CVRP)\cite{ nazari2018reinforcement, wu2021learning, li2023learning}, and the Team Orienteering Problem (TOP) \cite{sankaran2022gamma, wan2024deep}, which have demonstrated impressive performance in terms of solution quality and computational efficiency.

While the above DRL approaches have shown effectiveness in solving standard routing problems, a growing body of literature has focused on extending these methods to address various variants of routing problems in order to better align with the complexities of real-world applications. 
For instance, Wang et al. \cite{wang2024deep} proposed a neural heuristic based on DRL to solve the traditional and improved vehicle routing problem with backhauls (VRPB) variants. 
Li et al. \cite{Li2021DeepRL} designed a DRL-based algorithm to tackle heterogeneous CVRP (HCVRP), where vehicles are mainly characterized by different capacities.
In addition, other variant factors, such as time window constraints \cite{lin2021deep}, multiple depots \cite{arishi2023multi}, and dynamic traffic requests \cite{zhang2021solving}, have also been explored. By incorporating additional problem features and constraints, these studies have significantly broadened the applicability of DRL approaches in practical logistics and transportation scenarios.

In summary, the application of DRL methods to FSMVRP holds substantial academic and practical significance. While DRL-based methods have been successfully applied to a wide range of VRP variants, there remains a notable gap in the literature regarding their use for solving FSMVRP. Addressing this gap not only advances the theoretical understanding of DRL in complex routing problems but also provides valuable tools for efficient fleet management in modern logistics systems.

\section{Problem Statement} \label{problem_statement}
\begin{figure}
    \centering
    \includegraphics[width=0.9\columnwidth]{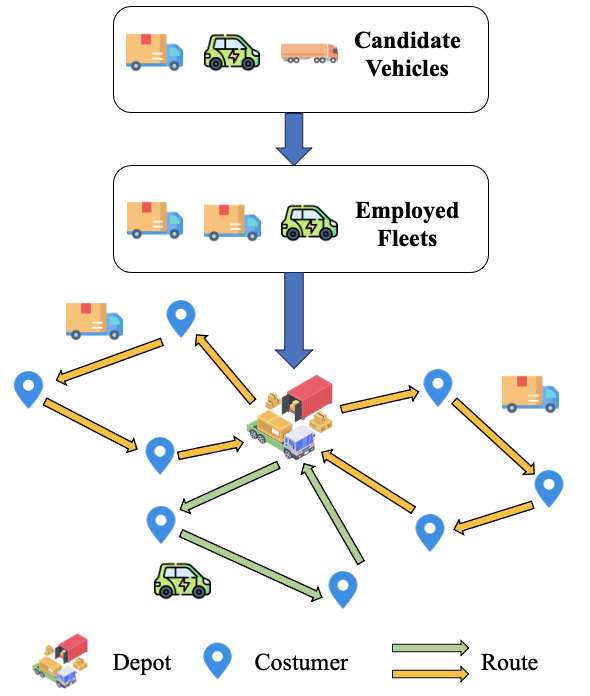}
    \caption{The example of the FSMVRP. Unlike the classic CVRP, FSMVRP requires consideration of both route planning and fleet composition. The inclusion of fleet composition introduces additional fixed costs, making it a critical factor that directly influences decision outcomes. Consequently, the need to optimize both routing and fleet structure renders FSMVRP more complex and challenging than the classic CVRP.} \label{fig1} 
\end{figure}

\noindent In this section, we formulate the standard mathematical model for FSMVRP first, then the corresponding MDP is built to show the decision-making process of DRL. The diagram depicting the FSMVRP is presented in Fig.\ref{fig1}.

\subsection{Mathematical Model} \label{model}

FSMVRP is usually defined on a complete graph $\mathcal{G} = (\mathcal{V}, \mathcal{E})$, where $\mathcal{V} = \mathcal{N}\cup \{0\}$ is the set comprising the depot (node 0) and a set of customers $\mathcal{N}=\{1,2,\dots, n\}$, and $\mathcal{E}$ is the set of edges connecting all nodes. A heterogeneous fleet of vehicles is available, comprising multiple types $k \in \mathcal{K}$, each characterized by a unique capacity $Q^{k}$, a fixed cost $f^{k}$ incurred per vehicle used, and a variable cost per unit distance $c^{k}$. Each customer $i \in \mathcal{N}$ has a deterministic demand $d_i$, and the travel distance between nodes $i$ and $j$ is denoted by $e_{ij}$. The objective of the FSMVRP is to minimize the total cost, which includes both the sum of the fixed costs for all vehicles utilized and the total variable travel costs, while determining a set of routes such that each customer is visited exactly once, each route starts and ends at the depot, and the total demand serviced by any vehicle of type $k$ does not exceed its capacity $Q^{k}$. The model is formulated as a mixed-integer linear program (MILP) with binary decision variables $x_{ij}^k$ indicating whether a vehicle of type $k$ traverses edge from $i$ to $j$, and continuous variables $u_i^k$ representing cumulative load delivered by a vehicle of type $k$ upon arrival at customer $i$. Then, the mathematical formulation of the FSMVRP is expressed as follows:

\begin{equation}
    \min \sum_{k \in \mathcal{K}}f^{k}\sum_{j \in \mathcal{N}}x_{0j}^{k} + \sum_{k \in \mathcal{K}}\sum_{i \in \mathcal{V}}\sum_{j \in \mathcal{V}, i \neq j}e_{ij} c^k x_{ij}^k 
\end{equation}
Subject to:
\begin{align}
    &\sum_{k \in \mathcal{K}}\sum_{i \in \mathcal{V}, i \neq j}x_{ij}^k = 1, \forall j\in \mathcal{N} \label{con_1} \\
    &\sum_{i\in \mathcal{N}}x_{i0}^k - \sum_{j\in \mathcal{N}}x_{0j}^k = 0, \forall k\in \mathcal{K} \label{con_2} \\
    &\sum_{i\in \mathcal{V}, i \neq j}x_{ij}^k - \sum_{i\in \mathcal{V}, i \neq j}x_{ji}^k = 0, \forall j \in \mathcal{N}, \forall k \in \mathcal{K} \label{con_3}\\
    &u_{i}^k - u_{j}^k + Q^{k}x_{ij}^k \leq Q^{k} - d_j x_{ij}^k, \forall i, j \in \mathcal{N}, i \neq j, \forall k \in \mathcal{K} \label{con_4}\\
    &d_i \sum_{j \in \mathcal{V}, i \neq j}x_{ji}^k \leq u_{i}^k \leq Q^{k}\sum_{j \in \mathcal{V}, i \neq j}x_{ji}^k, \forall i \in \mathcal{N}, \forall k \in \mathcal{K} \label{con_5}\\
    &x_{ij}^k \in \left\{0,1\right\} ,\forall i, j \in \mathcal{V}, i \neq j, \forall k\in \mathcal{K} \label{con_6}\\
    &u_{i}^{k} \geq 0, \forall i\in \mathcal{N}, \forall k \in \mathcal{K} \label{con_7}\\
    &u_{0}^k = 0, \forall k \in \mathcal{K} \label{con_8}
\end{align}
Constraint \eqref{con_1} guarantees that each customer is visited exactly once by a single vehicle.
Constraint \eqref{con_2} enforces flow conservation at the depot for each vehicle type, ensuring that the number of vehicles of each type departing from the depot equals the number returning.
Constraint \eqref{con_3} maintains flow balance at each customer node for every vehicle type, such that the number of vehicles entering a customer node equals the number leaving, thereby preserving route continuity.
Constraints \eqref{con_4} and \eqref{con_5} implement the Miller-Tucker-Zemlin (MTZ) subtour elimination conditions, which prevent the formation of infeasible sub-tours and ensure that the cumulative demand along any route does not exceed the vehicle’s capacity. 
Constraints \eqref{con_6}, \eqref{con_7}, and \eqref{con_8} define the ranges of variables $x_{ij}^k$ and $u_i^k$.

\subsection{Reformulation as MDP Form}

Since RL was initially introduced for tackling sequential decision-making challenges, the formulation of routes for the FSMVRP can similarly be interpreted as a sequential decision-making problem. In this work, we reformulate the decision-making process of FSMVRP with the MDP defined by a 4-tuple $\mathcal{M} = \{\mathcal{S}, \mathcal{A}, T, R\}$, which includes state space $\mathcal{S}$, action space $\mathcal{A}$, state transition $T$, and reward function $R$.

\textbf{State:} The state is composed of two main components. The first component captures node information, with a particular focus on unvisited customers, including their respective demands and locations. The second component represents fleet information, encompassing the current locations and loads of the vehicles in use, as well as the attributes of different vehicle types, such as fixed cost, unit travel cost (travel cost per unit distance), and capacity. Initially, no vehicles have been employed, and all vehicles are assumed to be located at the depot by default. Finally, all employed vehicles have to return to the depot.

\textbf{Action:} Different from some classic routing problems that only have to decide which node to visit at the current step \cite{kool2018attention}, we have to select both the vehicle and the node at each step. Specifically, the action at step $t$ is denoted as $a_t = (k^{j}_t,v^{i}_t)\in \mathcal{A}$, which means that vehicle of type $j$ visits node $i$ at step $t$. At each decision-making step, we only consider one vehicle for each type.
While FSMVRP necessitates deliberation on fleet configuration composition owing to fixed costs, we can determine the employment status of vehicles by evaluating their present states and actions. This approach helps circumvent the need for expanding the action space within the proposed decision framework, as elaborated in Section \ref{4.1}.

\textbf{Transition:} Since the FSMVRP is a static problem, the next state is fully determined by the current state and the selected action. Specifically, when a vehicle visits a customer, the location and load of the selected vehicle are updated accordingly, and the travel cost from the current position to the newly visited customer is incurred. Additionally, when a new vehicle is introduced into the fleet, the corresponding fixed cost is also incurred. 

\textbf{Reward:} The goal is to minimize the total cost associated with completing all tasks, where the total cost comprises both fixed and variable components. Since the proposed DRL-based approach is designed to maximize the objective function, the reward is defined as the negative of the total cost. This reward formulation effectively guides the policy towards solutions that achieve lower overall costs.
 
\section{Methodology} \label{method}

In this section, we detail our DRL-based approach for solving the FSMVRP. We begin by outlining the comprehensive framework of the proposed policy network FRIPN, which possesses the capability to compose fleet configurations and plan routes. Next, we describe the architecture of the policy network, including its encoder and decoder components. Finally, we present the training algorithm used to optimize the policy network. 

\subsection{Framework of Our Policy Network} \label{4.1}

In our proposed DRL-based approach, the objective is to learn a policy $\pi_{\theta}$ that can provide efficient decision-making under diverse demand scenarios and fleet employment conditions, where $\theta$ represents the trainable network parameters. Denote $\tau$ as the maximum number of decision steps, the joint probability of the whole decision process under policy $\pi_{\theta}$ can be reformulated by the chain rule:
\begin{equation}
    p(s_{\tau} | s_{0}) = \Pi_{t = 0}^{\tau -1} p(s_{t+1} | s_{t}, a_{t})\pi_{\theta}(a_t | s_t),
\end{equation}
where $p(s_{t+1} | s_{t}, a_{t}) = 1$ always holds since the next state is certain when actions are determined. Therefore, the results of FSMVRP totally depend on the effectiveness of the policy network.

\begin{figure}
    \centering
    \includegraphics[width=0.9\columnwidth]{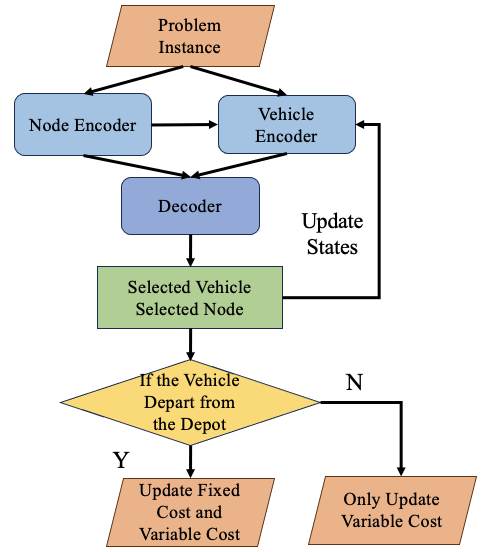}
    \caption{The framework of our policy network. The encoder-decoder framework decides the current vehicle and node. Whether a vehicle departs from the depot determines whether its fixed cost is included in the reward calculation.} \label{fig2} 
\end{figure}

The framework of our policy network FRIPN is illustrated in Fig. \ref{fig2}. To various problem instances, the policy network first builds an encoder part to record the information of the location distribution and the demand of nodes. Then, for each type, the current vehicle information will be encoded. Based on the output of the node encoder and the vehicle encoder, the policy network can select one vehicle and its next arrival node based on the encoded node graph and the current situation of all candidate vehicles. 
Specifically, regarding the selected vehicle: if it is currently located at the depot, this action corresponds to introducing a new vehicle into the fleet. Conversely, if the vehicle is not at the depot, it indicates that the vehicle has already been assigned for route planning. 
Through this network policy framework, we incorporate the fleet composition into the route scheduling process, thereby avoiding the potential high computational complexity caused by introducing additional fleet composition decision structures.

\subsection{Architecture of Our Policy Network} \label{4.2}

\begin{figure*}
    \centering
    \includegraphics[width=1.5\columnwidth]{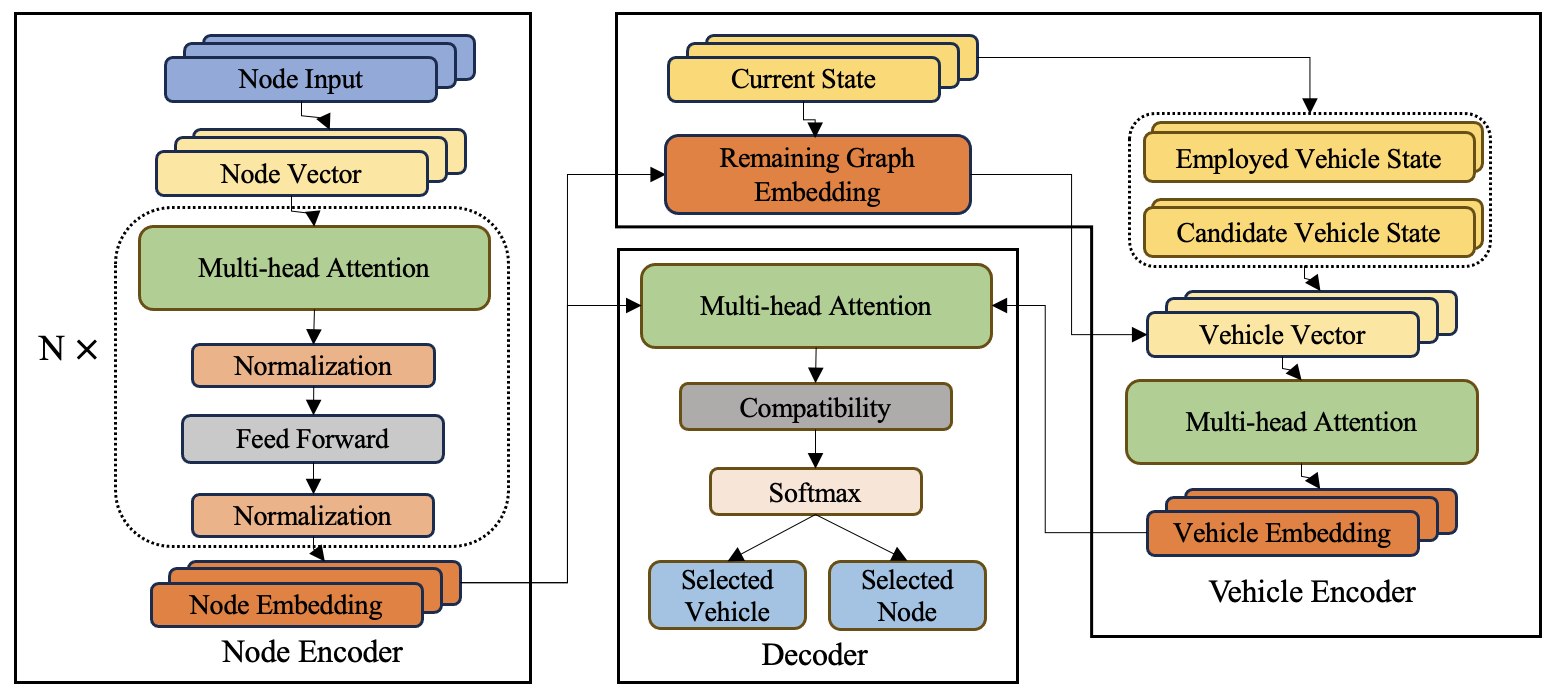}
    \caption{The architecture of our policy network. The policy network mainly includes the node encoder, vehicle encoder, and decoder.} \label{fig3} 
\end{figure*}

Based on the framework described in Section \ref{4.1}, we design the policy network architecture for FRIPN, as illustrated in Fig. \ref{fig3}. Similar to other DRL-based approaches for routing problems, our policy network adopts the Transformer architecture with an encoder-decoder structure. The encoder employs a multi-head attention (MHA) mechanism to generate node feature embeddings that capture high-dimensional relationships among nodes. The decoder then selects a vehicle and a node by querying these node embeddings using vehicle feature embeddings. A key challenge lies in constructing informative vehicle embeddings, as vehicles exhibit heterogeneity in fixed costs, variable costs, and capacities, factors that require explicit consideration. Our network is designed to effectively capture such vehicle heterogeneity, enabling it to employ appropriate vehicles and perform efficient route planning. The following subsections detail the node encoding, vehicle encoding, and decoding components. 

\subsubsection{Node Encoding}
The node encoding takes the nodes information as input, which consists of the depot location, customer locations, and customer demands. First, the information of each node is projected into a high-dimensional vector via a linear projection. These resulting vectors then serve as the input to the $L$-layer MHA-based structure, which subsequently generate the node embeddings. 
Specifically, each layer consists of a MHA sublayer and a feed-forward (FF) sublayer, and the input of all layers except for the first layer is the output of the previous layer. For example, denote the input of the $l$-th layer as $h_l$, and the number of head is $y$, then the attention value of head $i$ is:
\begin{equation}
    Z_{l,i} = \mathrm{softmax}\left(\frac{Q_{l,i}K_{l, i}{}^{\text{T}}}{\sqrt{dim_{k}}}\right)V_{l,i}, i \in \{1, 2, \cdots, y\},
\end{equation}
\begin{equation}\label{Qli}
    Q_{l,i} = h_{l}W^{Q}_{l,i}, K_{l,i} = h_{l}W^{K}_{l,i}, V_{l,i} = h_{l}W^{V}_{l,i},
\end{equation}
where $W^{Q}_{l,i}, W^{K}_{l,i}, W^{V}_{l,i}$ are learnable weight matrices for head $i$ in layer $l$, and $dim_{k}$ is the key dimension. Then the output of the $i$-th MHA sublayer is:
\begin{equation}
    \mathrm{MHA}(h_{l}) = \mathrm{Concat}(Z_{l,1}, Z_{l,2}, \cdots, Z_{l,y})W^{O}_{l}, \label{mha1}
\end{equation}
where $W^{O}_{l}$ is the output projection matrix.
With the skip-connection and instance normalization techniques, the normalized output is
\begin{equation}
    n_{l} = IN(\mathrm{MHA}(h_{l}) + h_{l}), \label{mha2}
\end{equation}
Similarly, given the FF sublayer, the output of $l$-th layer is 
\begin{equation}
    h_{l+1} = IN(FF(n_{l}) + n_{l}). \label{mha3}
\end{equation}
With the $L$-layer MHA-based structure, the fixed node embedding is obtained. Since the node information is static and to avoid additional computational complexity, the node encoding in one instance will only be performed once.

\begin{figure}
    \centering
    \includegraphics[width=1\columnwidth]{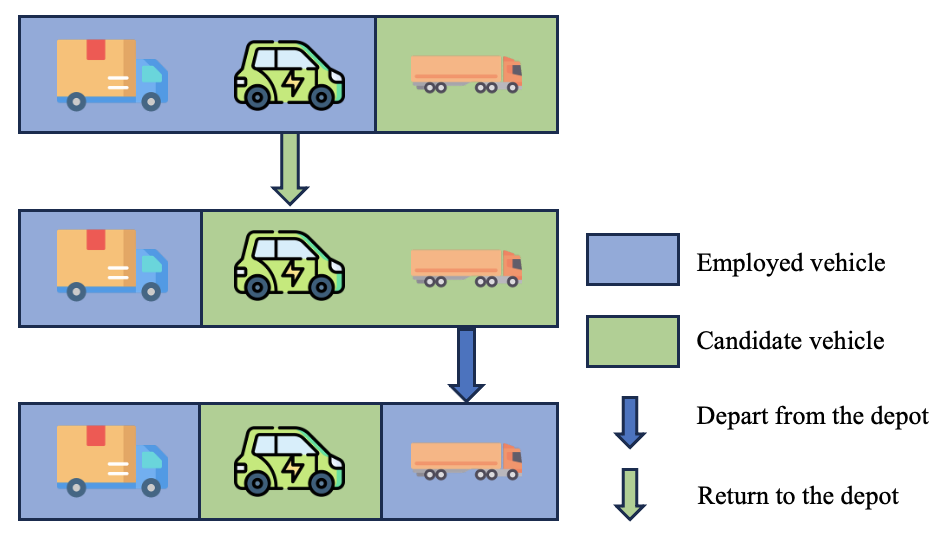}
    \caption{The transform between employed and candidate vehicles. The type of vehicle is determined by its current location, and it also affects the encoding of its operational state.} \label{fig4} 
\end{figure}

\subsubsection{Vehicle Encoding}

The vehicle encoding is designed to represent vehicle information at each step throughout the decision-making process. Unlike node encoding, the vehicle encoding is applied once per vehicle type at each step, to capture the current state of all vehicles that are potentially available for action. 
Specifically, vehicles of each type can be divided into employed vehicles and candidate vehicles, and the types they belong to can be transformed by actions, as shown in Fig. \ref{fig4}.
The input to the vehicle encoding consists of both fixed and variable attributes. The fixed attributes include the fixed cost, unit travel cost, and capacity for each vehicle type. The variable attributes comprise the current location and load of each vehicle. 

It is important to note that the total cost in FSMVRP comprises both the fixed cost of the fleet and the variable costs associated with travel. These distinct cost types must be handled differently when encoding vehicle states. Specifically, variable costs are incurred with each movement; hence, at every decision step, the unit travel cost of each vehicle must be reflected in the encoding. In contrast, fixed costs are only incurred when a vehicle leaves the depot (indicating its employment) and do not increase during subsequent travel of this vehicle. Consequently, the fixed cost is incorporated into the vehicle encoding exclusively at the point of depot departure. Let $AF_{1,t}^{k}$ denote the fixed cost assignment function for vehicle $k$ at step $t$, which is formally defined as follows:
\begin{equation}
    AF_{1,t}^{k}(f^k) = \left\{
    \begin{array}{ll}
    f^k & \begin{array}{@{}l@{}} \text{if vehicle $k$ is candidate vehicle} \\ \text{at step $t$,} \end{array} \\
    0 & \text {otherwise.}
    \end{array}\right.
\end{equation}
This formulation ensures that the fixed cost $f^k$ is incurred only when vehicle $k$ is the candidate vehicle, consistent with the typical fixed cost structure in FSMVRP. In all other states, the fixed cost contribution is zero.

In addition to vehicle fixed costs, our observations indicate that the presence of unvisited nodes plays a critical role in vehicle deployment decisions. Classical vehicle routing problems often disregard fixed costs, resulting in decoding processes that primarily evaluate selection probabilities to candidate nodes from the current state, while masking infeasible or previously visited nodes from the action space. However, when vehicle fixed costs have a significant impact on the overall objective, the decoding process must also account for the potential introduction of additional vehicles to serve the remaining unvisited nodes. Furthermore, the characteristics of the residual demand can influence not only the decision to deploy new vehicles but also the selection of appropriate vehicle types.
Let the final output of the node encoding be denoted as $h_{\text{NE}} = \{h_{\text{NE}}^0, h_{\text{NE}}^1, \dots, h_{\text{NE}}^n \}$. The remaining graph embedding for vehicle $k$ at step $t$ is defined as
\begin{equation}
    h_{\text{LG}, t}^{k} = \frac{\sum_{i \in N} \hat{h}_{\text{NE}, t}^{k, i}}{n+1},
\end{equation}
where
\begin{equation}
    \hat{h}_{\text{NE}, t}^{k, i} = \left\{
    \begin{array}{ll}
    h_{\text{NE}}^i & \begin{array}{@{}l@{}} \text{if node $i$ is available for vehicle $k$} \\ \text{at step $t$,} \end{array} \\
    0 & \text {otherwise.}
    \end{array}\right.
\end{equation}
Since the remaining graph embedding is designed to facilitate the decision of whether to introduce a new vehicle to complete the remaining tasks, it is considered only for candidate vehicles. Let $AF_{2, t}^{k}$ denote the remaining graph assignment function for vehicle $k$ at step $t$, which is formulated as
\begin{equation}
    AF_{2,t}^{k}(h_{\text{LG}, t}^{k}) = \left\{
    \begin{array}{ll}
    h_{\text{LG}, t}^{k} & \begin{array}{@{}l@{}} \text{if vehicle $k$ is candidate vehicle} \\ \text{at step $t$,} \end{array} \\
    0 & \text {otherwise.}
    \end{array}\right.
\end{equation}

After constructing the individual embeddings for each vehicle to represent their respective states, a MHA-based structure is employed to facilitate information sharing among the fleet. Let $h_{\text{IE}, t}$ denote the individual embeddings of all vehicles at step $t$. The aggregated vehicle embedding $h_{\text{VE}, t}$, which incorporates information from other vehicles in the fleet, is computed as follows: 
\begin{equation}
    h_{\text{VE}, t} = \mathrm{MHA}(h_{\text{IE}, t}W^{Q}_{v}, h_{\text{IE}, t}W^{K}_{v}, h_{\text{IE}, t}W^{V}_{v}),
\end{equation}
where $W^{Q}_{v}, W^{K}_{v}, W^{V}_{v}$ are learnable weight matrices corresponding to the query, key, and value projections, respectively. The computation procedure follows the same principles as described in equations \eqref{mha1}, \eqref{mha2}, and \eqref{mha3}.

\subsubsection{Decoding}

At decoding step $t$, the inputs consist of the fixed node encoding $h_{\text{NE}}$ and vehicle embedding $h_{\text{VE}, t}$, while the outputs are the selected vehicle and node for the current step.
To effectively aggregate vehicle state information with graph information, an MHA-based structure is also used. Unlike standard self-attention mechanisms, in this setup, the query is derived from the vehicle embedding, whereas the key and value are obtained from the node embedding. The aggregation is computed as follows:
\begin{equation}
    h_{c, t} = \mathrm{MHA}(h_{\text{VE}, t}W^{Q}_{c}, h_{\text{NE}, t}W^{K}_{c}, h_{\text{NE}, t}W^{V}_{c}),
\end{equation}
where $W^{Q}_{c}, W^{K}_{c}, W^{V}_{c}$ are learnable weight matrices.
Then the single-head attention mechanism is used to get the action selection probabilities. The query comes from the aggregation embedding $h_{c, t}$ and the key comes from the fixed node embedding $h_{\text{NE}}$. Subsequently, the computed embeddings are utilized to derive the query and key representations, as illustrated in \eqref{qst-eq} and \eqref{kst-eq}:
\begin{equation}\label{qst-eq}
    Q_{s, t} = h_{c, t}W^{Q}_{s}, Q_{s, t} = \{Q_{s, t}^{j}, j\in \mathcal{K}\}
\end{equation}
\begin{equation}\label{kst-eq}
    K_{s, t} = h_{\text{NE}}W^{Q}_{s}, K_{s, t} = \{K_{s, t}^{i}, i\in \mathcal{V}\}
\end{equation}
The attention weight between vehicle $j$ and node $i$ is therefore calculated as
\begin{equation}\label{uij}
    u_{i, j}^{t}=\left\{
    \begin{array}{ll}
    C \cdot \text{tanh} \left(\frac{Q_{s, t}^{j} {K_{s, t}^{i}}^{\text{T}}}{\sqrt{dim_{k}}} \right) & \begin{array}{@{}l@{}} \text{if node $i$ is available} \\ \text{for vehicle $j$,} \end{array} \\
    -\infty & \text {otherwise,}
    \end{array}\right.
\end{equation}
where $C$ denotes the clipping factor. By adjusting the compatibility between vehicles and nodes, actions that violate the constraints can be prohibited. The action probability is calculated by the \textit{softmax} function as follows:
\begin{equation}\label{sftm}
  \mathcal{P}\left(a_{i, j}^{t}\right)=\frac{e^{u_{i, j}^{t}}}{\sum_{j} \sum_{i} e^{u_{i, j}^{t}}}.
\end{equation}
The action at step $t$ is achieved by the action probability and the decoding strategies.

The decoding strategies include the greedy strategy and the sampling strategy. The former always selects the action with the highest probability at each step, while the latter samples solutions with a fixed size based on the action probabilities. In addition, we apply an instance augmentation technique that reformulates a given problem into multiple equivalent instances with the same solution \cite{kwon2020pomo}. By testing the model on these augmented instances, we can explore solution trajectories from different angles and aim to obtain better solutions.

\begin{algorithm}[htbp]
    \caption{Policy Training for FRIPN}
    \label{algorithm:reinforce}
    \KwIn {Numbers of epochs $E$; Number of training steps in one epoch $T$; Number of batches $B$; Number of trajectories for each instance $J$; Fleet size set $Z$;}
    Initialize the parameter of policy network $\theta$\;
    \For {epoch $= 1, 2, \dots, E$}{
        \For {step $= 1, 2, \dots, T$}{
            Sample the fleet size $z$ from $Z$\ randomly\;
            Sample $B$ instances based on $z$ randomly\;
            \For {batch $= 1, 2, \dots, B$}{
            Sample trajectories $\{\tau^{1}, \tau^{2}, \dots, \tau^{J}\}$ for instance\;
            Get the total cost for each trajectory\;
            Calculate the shared baseline according to formula \eqref{baseline}\; 
        }
        Update the parameter $\theta$ with gradient ascent $\nabla_{\theta} J(\theta)$\ in formula \eqref{update};
        }
    }
\end{algorithm}

\subsection{Policy Training}

The policy training algorithm, based on REINFORCE \cite{williams1992simple}, is presented in Algorithm \ref{algorithm:reinforce}. Due to the substantial variability in results across different instances, we employ a shared baseline rather than greedy rollout, exponential, or critic baselines \cite{bello2016neural, kool2018attention}. The shared baseline considers only the reward differences among trajectories generated from the same instance, thereby providing a more stable reference for policy updates \cite{kwon2020pomo}. Specifically, for each instance $i$, we sample $N$ solution trajectories $\{\tau^{1}_{i}, \tau^{2}_{i}, \cdots , \tau^{N}_{i} \}$, and compute the shared baseline as
\begin{equation}
    b_{i} = \frac{1}{N}\sum_{j=1}^{N}R(\tau^{j}_{i}). \label{baseline}
\end{equation}
where $R(\tau^{j}_{i})$ denotes the reward of trajectory 
$\tau^{j}_{i}$. Using this baseline, the policy parameters $\theta$ are optimized via gradient ascent: 
\begin{equation}
    \nabla_{\theta} J_{i}(\theta) \approx \frac{1}{N}\sum_{j=1}^{N}(R(\tau^{j}_{i})-b_{i})\nabla_{\theta} {\rm log} p_{\theta}(\tau^{j}_{i}), \label{update}
\end{equation}
where $J_{i}(\theta)$ represents the expected return for instance $i$. Furthermore, since the number of vehicle categories varies across instances, we randomly sample the scale of vehicle categories prior to each batch update, and then generate the training data for that batch based on the sampled parameter. This approach ensures that the scale of vehicle categories remains consistent within each batch, thereby mitigating the impact of these differences on batch learning efficiency.

\section{Experiments} \label{experiment}

In this section, we assess the efficacy of the proposed DRL-based approach on FSMVRP. Initially, we perform ablation experiments to demonstrate the impact of the remaining graph embedding. Subsequently, we contrast our method with conventional baselines across problems of varying scales. Finally, we validate the effectiveness of our method by testing it on benchmark instances. Our code is available online at \href{https://github.com/PFW-coder/FSMVRP-DRL.git}{https://github.com/PFW-coder/FSMVRP-DRL.git} 

\subsection{Experimental Settings}

\subsubsection{Datasets}\label{datasets}

Our experiments are conducted on two types of datasets: randomly generated datasets and benchmark datasets. For the randomly generated datasets, we consider problem instances of varying scales, with the number of customer nodes set to 20, 50, 75, and 100. For each instance, the number of vehicle types ranges from 3 to 6. Following established practices in the literature \cite{kool2018attention, wang2024deep}, the coordinates of both customer nodes and the depot are uniformly sampled from the square region $[0, 1] \times [0, 1]$. Customer demands are uniformly generated from the interval $[0.01, 0.5]$, and vehicle capacities are uniformly sampled from $[0.5, 3]$. 
Since the fixed cost is typically related to vehicle capacity, we first randomly generate a standard linear factor for the fixed cost. Specifically, for each instance, the standard linear factor  $f_{sta}$ is uniformly sampled from $[1, 20]$. The specific linear factor for each vehicle type is then generated from $[f_{sta} - 1, f_{sta} + 1]$, ensuring that these values fluctuate around $f_{sta}$. To maintain validity, the specific linear factors are clamped to a minimum value of 1. Denoting the specific linear factor for vehicle $k$ as $f_{sta}^k$, the fixed cost for vehicle $k$ is calculated as $f_{sta}^k Q^k$. The variable cost for each vehicle is uniformly sampled from $[1, 3]$. 
For the benchmark dataset, we utilize 12 standard FSMVRP instances \cite{golden1984fleet}. In these instances, the fixed costs vary among candidate vehicles, while the variable costs remain constant. Since our DRL-based approach is designed for scenarios where both fixed and variable costs are considered for candidate vehicles, we exclude special cases that do not involve fixed costs.

\subsubsection{Training and Testing}

For each problem, we train the policy network for 2000 epochs. In each epoch, the batch size is 32 and the training episode is 10000. The initial learning rate is 0.0001 and the optimizer is Adam. During the training process, we generate random instances for training.
For testing, we used two datasets mentioned in Section \ref{datasets}. For the randomly generated datasets, we generate 1000 instances by the fixed random seed for each problem. Our proposed method and baselines are all used to test the randomly generated datasets. The maximum runtime allowed for each baseline on every instance is set to 1800 seconds. For the benchmark datasets, we directly use the best results of the existing literature. 

\subsubsection{Baselines}

We compare the proposed DRL-based method with various baselines, including the exact solver and heuristic-based methods. Baselines introduced here are applied to test the randomly generated datasets, and the related details are described as follows.

\begin{itemize}
    \item \textbf{CPLEX:} CPLEX is widely adopted for linear programming models, which can get exact optimal results. It has been successfully applied in various routing problems for benchmarks \cite{wang2024deep}. The mathematical model for CPLEX solver is displayed in Section \ref{model}.
    \item \textbf{Adaptive Large Neighborhood Search:} ALNS is a powerful heuristic-based method to solve various routing problems by iteratively destroying and repairing parts of the solution using a set of adaptive operators \cite{hiermann2016electric, masmoudi2022fleet}. ALNS operates by repeatedly selecting and applying different destroy and repair operators to modify the current solution, where the choice of operators is guided by an adaptive mechanism based on their historical performance.
    \item \textbf{Tabu Search:} TS is a local search heuristic that explores the solution space by iteratively moving to the best neighboring solution while avoiding cycles through the use of a tabu list \cite{glover1989tabu}. It efficiently escapes local optima by forbidding or penalizing recently visited solutions for a certain number of iterations. Tabu Search is chosen as a baseline due to its effectiveness and reliability in solving various VRP variants, including FSMVRP \cite{brandao2009deterministic}.
    \item \textbf{Genetic Algorithm:} GA is a population-based metaheuristic that evolves solutions through iterative processes of selection, crossover, and mutation inspired by natural evolution \cite{holland1992genetic}. It maintains diversity in the solution pool and efficiently explores the search space by combining and modifying candidate solutions. GA has been widely used to solve various VRP variants \cite{liu2009effective}.
    \item \textbf{Compared DRL:} Due to the lack of existing DRL-based methods for FSMVRP, we design the compared DRL-based method. Compared to our proposed method, we do not apply the remaining graph embedding to assist in fleet composition. 
\end{itemize}

\subsection{Ablation Experiments}

\begin{figure}
    \centering
    \includegraphics[width=1\columnwidth]{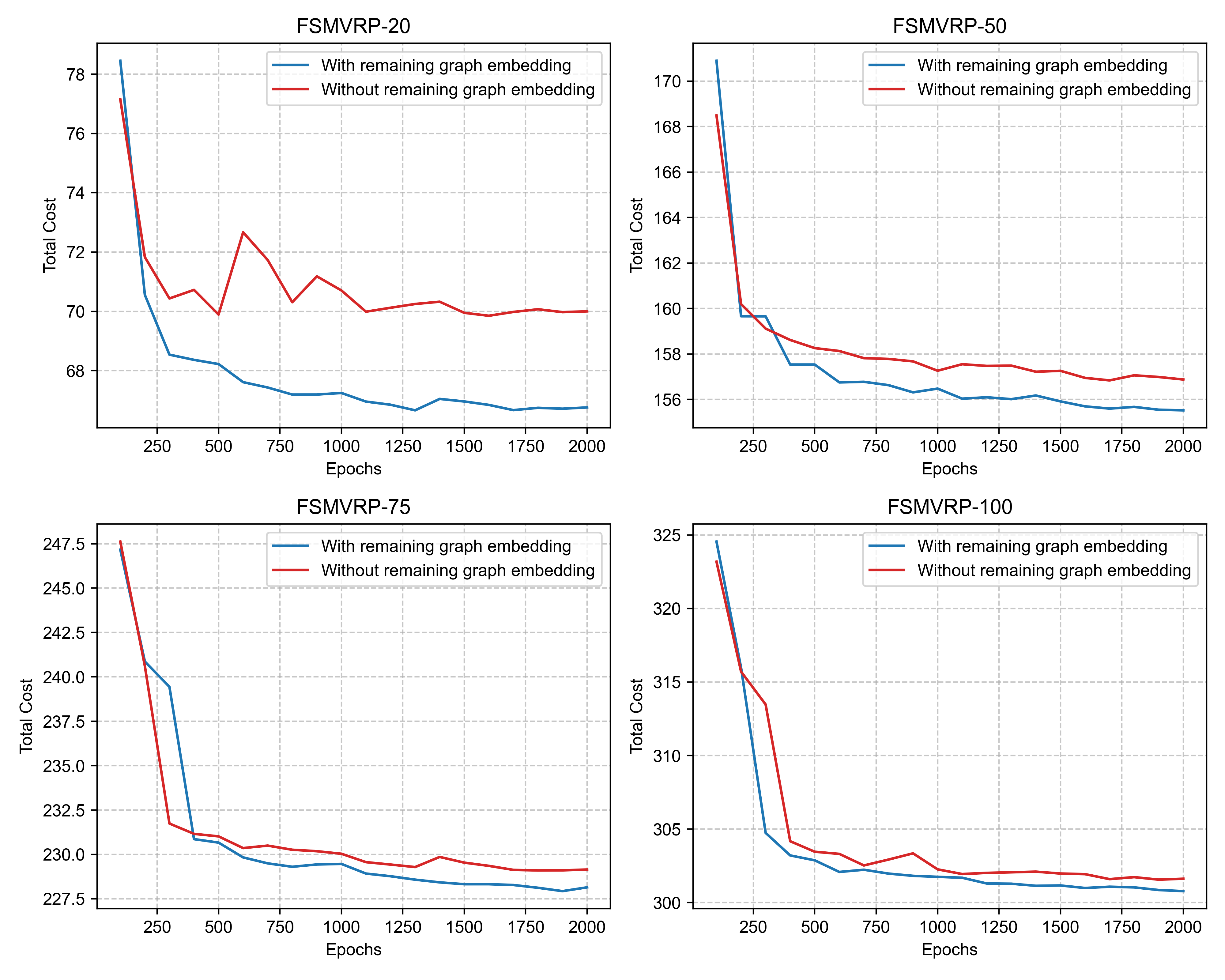}
    \caption{Ablation experiment results. Regardless of problem size, incorporating the remaining graph embedding consistently facilitates convergence to superior results. } \label{fig5} 
\end{figure}

Given that a key characteristic of FSMVRP is the composition of candidate fleets for route planning, we have incorporated remaining graph embeddings to enhance the vehicle employment process. To rigorously evaluate the contribution of remaining graph embeddings to our approach, it is essential to conduct ablation experiments. We train policy networks with and without remaining graph embedding separately, and output their training curves under greedy encoding. The results of the ablation study are presented in Fig. \ref{fig5}. By testing problems of various scales, the remaining graph embeddings always lead to better convergence results. These experiments provide critical insights into the effectiveness and necessity of remaining graph embeddings within our framework.

\subsection{Comparison Study} \label{comparison-study}

The comparison results are summarized in Table \ref{table_1}. For each method, we report the average cost, runtime, and cost gap across all test instances. The cost gap is calculated by using the lowest cost obtained for each problem as the benchmark and computing the corresponding percentage gap for each method. For DRL-based methods, we consider various combinations of decoding strategies, and present the results for each combination separately. Both the sampling strategy and the instance augmentation technique are shown to contribute to improved performance for DRL-based methods.
Regarding CPLEX, only the results for instances with 20 nodes are reported, as its performance deteriorates significantly on larger-scale instances. Moreover, CPLEX often fails to find optimal solutions for 20-node instances within the 1800-second time limit. The comparison indicates that heuristic-based methods, such as ALNS and TS, perform well on small-scale instances. For example, ALNS and TS achieve better results than our proposed DRL-based method on instances with 20 nodes, with our method incurring $1.74\%$ higher cost. However, as the instance size increases, our DRL method demonstrates increasing advantages. Specifically, on instances with 100 nodes, our method outperforms the best heuristic-based method by $8.80\%$. Furthermore, our proposed DRL method consistently outperforms the compared DRL baseline under the same decoding strategy across various instance scales, which aligns with the findings from our ablation experiments.
It is also worth noting that DRL-based methods can obtain solutions within a few seconds, offering significant advantages in time-constrained scenarios.

\begin{table*}[htp]
	\centering
	\caption {Comparison results on randomly generated instances}
	\label{table_1}
	\begin{tabular}[c]{*{13}{c}}
            \toprule
            \multirow{2}{*}{Method} & \multicolumn{3}{c}{FSMVRP-20} & \multicolumn{3}{c}{FSMVRP-50} & \multicolumn{3}{c}{FSMVRP-75} & \multicolumn{3}{c}{FSMVRP-100} \\
		\cmidrule(lr){2-4}
            \cmidrule(lr){5-7}
            \cmidrule(lr){8-10}
            \cmidrule(lr){11-13}
            \multicolumn{1}{c}{} & \multicolumn{1}{c}{Cost} & \multicolumn{1}{c}{Time (s)} & \multicolumn{1}{c}{Gap} & \multicolumn{1}{c}{Cost} & \multicolumn{1}{c}{Time (s)} & \multicolumn{1}{c}{Gap} & \multicolumn{1}{c}{Cost} & \multicolumn{1}{c}{Time (s)} & \multicolumn{1}{c}{Gap} & \multicolumn{1}{c}{Cost} & \multicolumn{1}{c}{Time (s)} & \multicolumn{1}{c}{Gap} \\
		\midrule
		  CPLEX & 78.28 & 1800 & 25.08\% & N/A & N/A & N/A & N/A & N/A & N/A & N/A & N/A & N/A\\
            ALNS & \textbf{62.96} & 1800 & 0.00\% & 155.77 & 1800 & 2.83\% & 239.88 & 1800 & 7.45\% & 321.30 & 1800 & 8.80\%\\
            TS & 63.79 & 1800 & 1.33\% & 161.66 & 1800 & 6.37\% & 246.55 & 1800 & 10.43\% & 328.14 & 1800 & 11.11\%\\
            GA & 66.34 & 1800 & 5.36\% & 166.12 & 1800 & 9.75\% & 253.33 & 1800 & 13.47\% & 340.19 & 1800 & 15.20\%\\
            \midrule
            \makecell[c]{Compared DRL \\ (Greedy, no augment)}
            & 69.94 & 0.87 & 11.08\% & 157.12 & 0.98 & 3.80\% & 229.02 & 1.11 & 2.58\% & 301.63 & 1.17 & 2.14\%\\
            \makecell[c]{Compared DRL \\ (Greedy, augment)} & 66.60 & 0.86 & 5.78\% & 154.57 & 0.98 & 2.12\% & 226.56 & 1.11 & 1.48\% & 298.91 & 1.19 & 1.22\%\\
            \makecell[c]{Compared DRL \\ (Sampling, no augment)} & 65.14 & 0.90 & 3.47\% & 152.51 & 1.03 & 0.76\% & 224.52 & 1.44 & 0.57\% & 296.64 & 1.48 & 0.45\% \\
            \makecell[c]{Compared DRL \\ (Sampling, augment)} & 64.32 & 0.92 & 2.16\% & 151.91 & 1.22 & 0.36\% & 223.60 & 1.47 & 0.16\% & 295.97 & 1.73 & 0.22\%\\
            \midrule
            \makecell[c]{Ours \\ (Greedy, no augment)} & 66.66 & 0.88 & 5.88\% & 155.54 & 0.98 & 2.76\% & 227.84 & 1.09 & 2.05\% & 300.58 & 1.19 & 1.78\%\\ 
            \makecell[c]{Ours \\ (Greedy, augment)} & 65.27 & 0.88 & 3.67\% & 153.50 & 1.01 & 1.41\% & 225.64 & 1.10 & 1.07\% & 298.13 & 1.19 & 0.95\%\\
            \makecell[c]{Ours \\ (Sampling, no augment)} & 64.55 & 0.91 & 2.53\% & 152.01 & 1.06 & 0.43\% & 224.13 & 1.27 & 0.39\% & 296.22 & 1.36 & 0.31\%\\
            \makecell[c]{Ours \\ (Sampling, augment)} & 64.05 & 1.09 & 1.74\% & \textbf{151.36} & 1.35 & 0.00\% & \textbf{223.25} & 1.94 & 0.00\% & \textbf{295.32} & 2.64 & 0.00\%\\
		\bottomrule
	\end{tabular}
\end{table*}

\subsection{Evaluation on Benchmark Dataset}

Since the benchmark dataset proposed by Golden et al. \cite{golden1984fleet} has been widely adopted in the literature, we evaluate our proposed DRL method on this dataset for a comprehensive comparison. Specifically, instances 3–6 contain 20 nodes, instances 13–16 contain 50 nodes, instances 17 and 18 contain 75 nodes, and instances 19 and 20 contain 100 nodes.
It is important to note that the training data configuration for our method differs from the benchmark dataset settings. First, while the benchmark dataset assumes uniform variable costs across different vehicle types, our training data setting incorporates distinct variable costs for each vehicle type. Second, the vehicle parameter settings in the benchmark instances, such as the fixed cost, may fall outside the range of our training data. Despite these discrepancies, we do not retrain our policy for these specific instances; instead, we directly evaluate the pre-trained policy, thereby demonstrating the generalization capability of our approach and highlighting its potential applicability to real-world scenarios.
In addition, we employ a decoding strategy that combines sampling and instance augmentation, which, as analyzed in Section \ref{comparison-study}, yields the best performance. The experimental results are summarized in Table \ref{table_2}. Our learned policy achieves near-optimal solutions on the benchmark instances, with an average cost gap of only 1.10\% compared to the results of best-known algorithms. Furthermore, our method consistently generates solutions within a few seconds, outperforming other algorithms, particularly on larger-scale instances. 

\begin{table*}[htp]
	\centering
	\caption {Comparison results on benchmark dataset}
	\label{table_2}
    \resizebox{\linewidth}{!}{
	\begin{tabular}[c]{*{16}{c}}
            \toprule
            \multirow{3}{*}{Instance} & \multicolumn{2}{c}{\makecell[c]{Taillard \\ (1999)\cite{taillard1999heuristic}}} & \multicolumn{2}{c}{\makecell[c]{Gendreau et al. \\ (1999)\cite{gendreau1999tabu}}} & \multicolumn{2}{c}{\makecell[c]{Lima et al. \\ (2004)\cite{lima2004memetic}}} & \multicolumn{2}{c}{\makecell[c]{Choi and Tcha \\ (2007)\cite{choi2007column}}} & \multicolumn{2}{c}{\makecell[c]{Liu et al. \\ (2009)\cite{liu2009effective}}} & \multicolumn{2}{c}{\makecell[c]{Guezouli and \\ Abdelhamid (2017)\cite{guezouli2017multi} }} & \multicolumn{3}{c}{Ours} \\
		\cmidrule(lr){2-3}
            \cmidrule(lr){4-5}
            \cmidrule(lr){6-7}
            \cmidrule(lr){8-9}
            \cmidrule(lr){10-11}
            \cmidrule(lr){12-13}
            \cmidrule(lr){14-16}
            \multicolumn{1}{c}{} & \multicolumn{1}{c}{Cost} & \multicolumn{1}{c}{Time (s)} & \multicolumn{1}{c}{Cost} & \multicolumn{1}{c}{Time (s)} & \multicolumn{1}{c}{Cost} & \multicolumn{1}{c}{Time (s)} & \multicolumn{1}{c}{Cost} & \multicolumn{1}{c}{Time (s)} & \multicolumn{1}{c}{Cost} & \multicolumn{1}{c}{Time (s)} & \multicolumn{1}{c}{Cost} & \multicolumn{1}{c}{Time (s)} & \multicolumn{1}{c}{Cost} & \multicolumn{1}{c}{Time (s)} & \multicolumn{1}{c}{Gap}\\
		\midrule
		  3 & \textbf{961.03} & N/A & \textbf{961.03} & 164 & 963.40 & 89 & \textbf{961.03} & 0 & \textbf{961.03} & 0 & \textbf{961.03} & 0 & 979.22 & 1.06 & 1.89\%\\
            4 & \textbf{6437.33} & N/A & 6441.01 & 253 & \textbf{6437.33} & 85 & 6441.01 & 1 & \textbf{6437.33} & 0 & 6438.73 & 0 & 6442.22 & 0.98 & 0.08\%\\
            5 & 1008.59 & N/A & 1008.72 & 164 & 1009.90 & 85 & \textbf{1007.05} & 1 & 1007.20 & 2 & 1008.20 & 3 & 1008.59 & 1.06 & 0.15\%\\
            6 & \textbf{6516.47} & N/A & 6517.98 & 309 & \textbf{6516.47} & 85 & 6516.84 & 0 & \textbf{6516.47} & 0 & 6531.47 & 5 & 6518.13 & 0.95 & 0.03\%\\
            13 & 2436.78 & 470 & 2424.88 & 724 & 2409.10 & 559 & 2409.77 & 10 & 2414.50 & 91 & \textbf{2401.50} & 22 & 2496.12 & 1.91 & 3.94\%\\
            14 & 9123.60 & 570 & 9121.98 & 1033 & 9121.62 & 669 & \textbf{9119.13} & 51 & 9119.48 & 42 & 9121.48 & 45 & 9137.82 & 1.26 & 0.20\%\\
            15 & 2593.61 & 334 & 2590.68 & 901 & 2590.20 & 554 & 2588.92 & 10 & \textbf{2586.37} & 48 & 2591.37 & 68 & 2644.28 & 1.35 & 2.24\%\\
            16 & 2744.25 & 349 & 2743.96 & 815 & 2729.60 & 507 & 2731.08 & 11 & 2735.54 & 107 & \textbf{2723.54} & 110 & 2774.28 & 1.39 & 1.86\%\\
            17 & 1753.74 & 2072 & 1752.29 & 1022 & 1770.30 & 1517 & 1746.26 & 207 & \textbf{1745.54} & 109 & 1747.54 & 99 & 1786.66 & 1.95 & 2.36\%\\ 
            18 & 2382.80 & 2744 & 2392.57 & 691 & 2401.00 & 1613 & \textbf{2375.78} & 70 & 2377.92 & 197 & 2376.19 & 179 & 2457.30 & 2.54 & 3.43\%\\
            19 & 8665.40 & 12528 & 8682.50 & 1687 & 8695.30 & 2900 & \textbf{8665.08} & 1179 & 8666.04 & 778 & 8667.06 & 678 & 8760.59 & 2.40 & 1.10\%\\
            20 & 4063.18 & 2117 & 4100.20 & 1421 & 4142.30 & 2383 & 4057.33 & 264 & 4052.81 & 1004 & \textbf{4049.53} & 1021 & 4145.65 & 2.42 & 2.37\%\\
            Average & 4057.23 & 2648 & 4061.48 & 765 & 4065.54 & 920 & 4051.60 & 150 & 4051.69 & 198 & \textbf{4051.47} & 186 & 4095.91 & 1.61 & 1.10\%\\
		\bottomrule
	\end{tabular}}
\end{table*}

\subsection{Test in larger-scale scenarios}

\begin{figure}
    \centering
    \includegraphics[width=1\columnwidth]{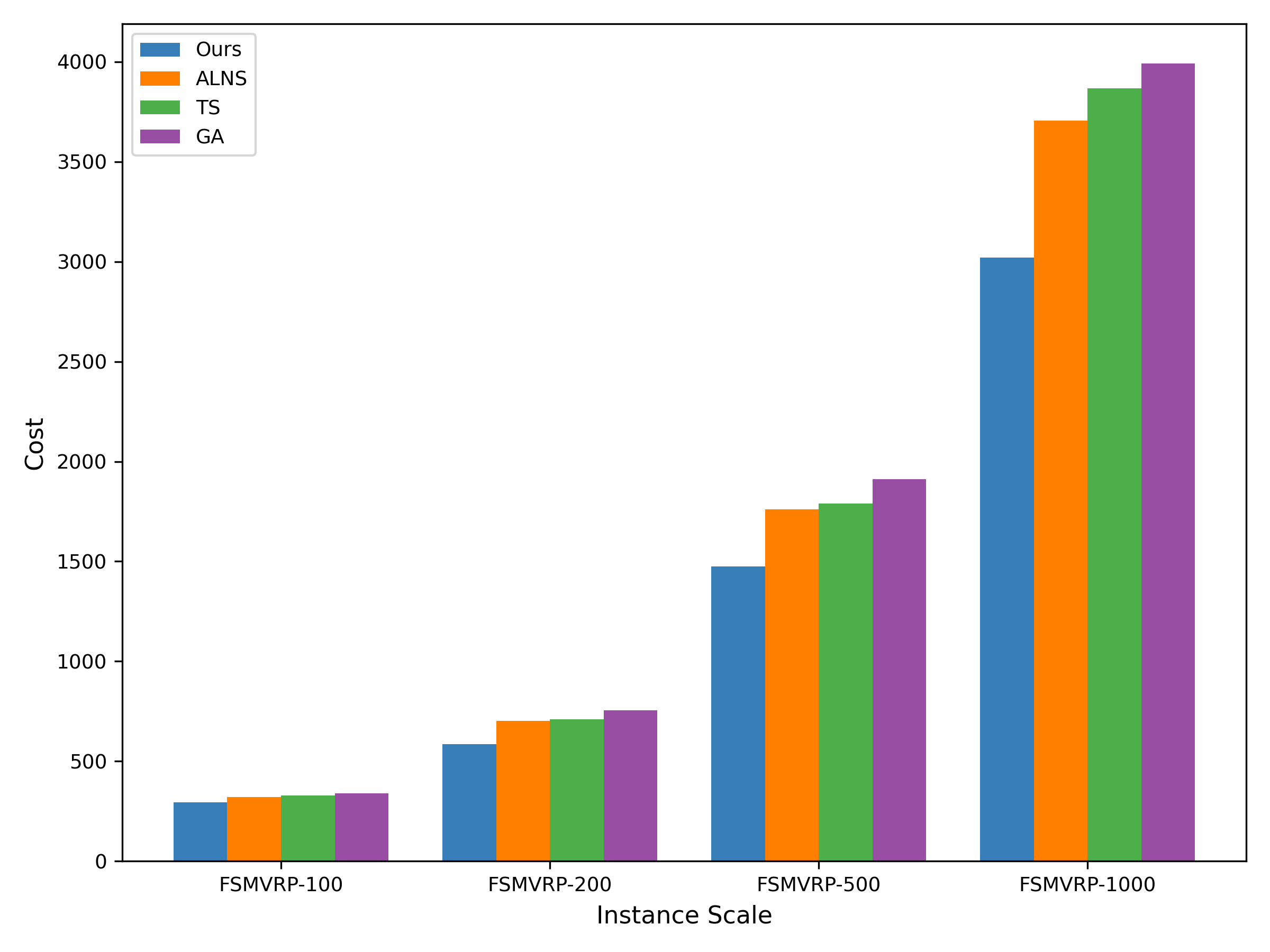}
    \caption{Comparison results on larger-scale instances. The advantages of our proposed DRL-based approach become more pronounced as the problem size increases.} \label{fig6} 
\end{figure}

To further evaluate the scalability of our proposed method, we conduct additional experiments on large-scale instances with 200, 500, and 1000 nodes. For these larger instances, we apply the policies learned from 100-node cases, rather than retraining new policies. The comparative results, presented in Fig. \ref{fig6}, demonstrate that our method consistently outperforms heuristic-based approaches as the problem size increases. Moreover, these findings highlight the strong generalization capability of our proposed method to larger and more complex scenarios.

\section{Conclusion and Future Works} \label{conclusion}

In this paper, we propose a DRL-based approach to address the FSMVRP, a significant variant of the classical VRP. To develop our algorithm, we formulate the problem as a MDP and design a corresponding policy network architecture. In particular, we introduce specialized input embeddings tailored to the unique requirements of fleet composition and routing decisions, including a remaining graph embedding that facilitates effective vehicle employment decisions. Our method is evaluated on both randomly generated and benchmark instances, demonstrating superior performance compared to existing approaches, especially in large-scale and time-constrained scenarios. For future work, further exploration of techniques related to vehicle employment decisions and policy training for CO is warranted. Additionally, DRL-based methods can be extended to address other variants of FSMVRP.

\bibliographystyle{ieeetr}
\bibliography{references}

\end{document}